# Assessment of texture measures susceptibility to noise in conventional and contrast enhanced computed tomography lung tumour images

**Omar Sultan Al-Kadi***
University of Sussex
Brighton BN1 9QH United Kingdom

**Abstract:** Noise is one of the major problems that hinder an effective texture analysis of disease in medical images, which may cause variability in the reported diagnosis. In this paper seven texture measurement methods (two wavelet, two model and three statistical based) were applied to investigate their susceptibility to subtle noise caused by acquisition and reconstruction deficiencies in computed tomography (CT) images. Features of lung tumours were extracted from two different conventional and contrast enhanced CT image data-sets under filtered and noisy conditions. When measuring the noise in the background open-air region of the analysed CT images, noise of Gaussian and Rayleigh distributions with varying mean and variance was encountered, and Fishers' distance was used to differentiate between an original extracted lung tumour region of interest (ROI) with the filtered and noisy reconstructed versions. It was determined that the wavelet packet (WP) and fractal dimension measures were the least affected, while the Gaussian Markov random field, run-length and co-occurrence matrices were the most affected by noise. Depending on the selected ROI size, it was concluded that texture measures with fewer extracted features can decrease susceptibility to noise, with the WP and the Gabor filter having a stable performance in both filtered and noisy CT versions and for both data-sets. Knowing how robust each texture measure under noise presence is can assist physicians using an automated lung texture classification system in choosing the appropriate feature extraction algorithm for a more accurate diagnosis.

*Keywords: texture analysis, feature extraction, CT image noise, contrast enhanced CT, lung tumour*

## 1 Introduction

Texture in computed tomography (CT) images can offer an important source of information on the state of the health of an examined organ. Diseased tissue usually has more rough or chaotic structure than the healthy counterparts, which can be characterised quantitatively for an automated diagnostic support system. The quality of the extracted texture measures is of significant importance for a correct diagnosis, especially when the difference between two different tissues becomes minor.

CT images are not immune to noise, and this can reduce the diagnostic quality of image texture. The resulting noise in the filtered backprojection (FBP) reconstructed CT images is usually homogenously distributed within the image field of view when compared to other iterative techniques which can provide an alternative way to reconstruct the raw data (i.e. sonogram), and significant blurring can occur due to not prefiltering when applying the FBP reconstruction algorithm. However, modern CT scanners deploy different filtering approaches which can reduce FBP reconstruction noise to unobservable levels. Yet, CT images can still be degraded by different levels of noise magnitude, which might arise for example, but is not limited to, from fluctuations in X-ray photons, low radiation doses, instability or deficiencies in the detectors' electronics receiver system and/or quantisation errors (i.e. due to rounding or truncation) [1], or even from the heart beating motion during the scanning time which could introduce subtle sampling artefacts despite the breath hold situation. These distortions affecting the fine structure of the examined tissue texture may obscure some prominent characteristics that distinguish one tumour subtype from another, or could decrease the tumour staging accuracy, and hence have a negative impact on the overall patients' prognosis. Therefore, having clear and relatively noise-free acquired images plays a significant role in medical image analysis. A number of studies applied various techniques in an endeavour to deal with noise issues in CT images, as reducing direct noise and streak artefacts [2-4], or removing statistical random noise [5-7]. Although all denoising techniques report a reduction in measured noise levels and better visual improvement, yet a complete removal of noise is not guaranteed and noise removal might be accompanied with a slight deformation or edge blurring of the tissue structure, reducing the differences between the various examined tissues. Usually tumour regions occupy a small portion of the acquired CT image, and the analysis is constrained to this small region of the diseased tissue for deriving discriminative features; whereas additive noise would further complicate the diagnostic process. Hence comes the importance of the applied texture measures to effectively characterise the tumour texture and how robust is their performance under noisy conditions (i.e. unfiltered images) or even in cleaned (filtered) images with remaining noise residues.

Physicians tend to use computed texture measures from regions of interest (ROIs) for diagnosis purposes and for eventually choosing the appropriate treatment procedure. Many techniques have been applied for the purpose of lungs texture analysis: as using the fractal dimension to exploit the fractal nature of the lung tissue structure [8-10], overcomplete wavelet filters – also called wavelet frames – to investigate the tissue at multiple resolutions [11, 12], combining Gabor filter response with histogram features [13], and using the co-occurrence matrix [14]. A review on the various methods used in computer analysis of lung CT scans can be found in [15]. However, we need to take into consideration when examining the texture of a small ROI in a medical image, that noise could adversely affect the accuracy of the measured texture parameters and cause errors in the reported diagnosis [16]. Although many studies concerned with noise reduction and CT image enhancement have been taken [2-7], yet there is a need to evaluate the texture measures' feature extraction performance under actual noisy conditions. The impact of additive white noise on Gabor filters and co-occurrence matrices, and on local power and phase spectra feature extraction ability from ordinary texture images of the Brodatz album was studied in [17, 18]; nevertheless, there is limited research in the literature regarding evaluating texture measures' performance under noisy conditions for medical CT images. This paper extends our earlier work [19] and aims to provide a comparison between seven different well-known texture measures to investigate their susceptibility to uncorrelated noise occurring in CT images, giving an indication of their reliability and fidelity in analysing lungs texture. The main emphasis was given to subtle statistical random noise rather than artefact noise which might appear as obvious streaks in the image.

In Section 2 we describe the procedure of noise estimation in each of the acquired CT images, and then how different texture features are extracted from the newly generated images given the estimated noise. This will be followed by measuring the separability quality of the applied texture measures. The results are then analysed and discussed in detail in Section 3, and the paper ends with a conclusion in Section 4.

**2 Material and Methods**

The type of noise needs to be first identified, and then two images are generated from each original CT image, one with a reduced noise (filtered image) and another with an enhanced noise (distorted image). These versions are CT reconstructed and two new ROIs ─ one from each of the two reconstructed versions ─ are extracted from the tumour area (Fig. 1a and 1d) and compared with the original ROI according to seven different texture measures. This process, which is summarised in Fig. 2, is iterated for all data-set images, while the used procedure is described in detail as in sub-sections 2.1-2.5.

2.1 Image acquisition

Two different data-sets of lungs infected with tumours of varying stages were available for analysis. The first data-set which was earlier used in our preliminary research [19], was a contrast enhanced (CE) CT angiography DICOM (Digital Imaging and Communication in Medicine) images referring to 11 patients (6 males and 5 females with age 63 ± 8 years old with lung cancer greater than 10 $mm^2$), having a resolution of 12 bits per pixel (bpp). The CE images were acquired using an Elscint CT-Twin scanner ─ after injecting a dose of an iodine contrast agent into one of the large veins of each patient ─ with X-ray tube voltage and current of 140 kV and 200mAs, a 10 mm slice thickness with matrix size 512 x 512, field of view (FOV) 430mm, 0° gantry angulation and B reconstruction filter.

The second data-set consists of conventional or non-contrast enhanced (NCE) CT images of 56 different cases of patients (31 males and 25 females with age 68 ± 10 years old) diagnosed also with lung cancer. A GE lightSpeed Ultra scanner was used to acquire the NCE CT images, while the acquisition parameters of the NCE images were similar to the CE data-set with the only difference in the resolution and slice thickness, where the NEC CT had an improved resolution of 16 bpp and a thinner slice thickness of 2mm. In both data-sets, patients were asked to hold their breath, if possible, or at least to breathe quietly during image acquisition in order to reduce artifactual distortions. It should be stated that all acquired images were ethically approved, and our work did not influence the diagnostic process or the patient's treatment.

2.2 Noise evaluation

The original image is first inspected for presence of noise, and the type of noise is appropriately identified for removal without destroying the fine structure of the image texture. Two new images will be produced from this phase, a clean (i.e. filtered original image) and distorted (i.e. the detected noise in the original image is doubled) versions.

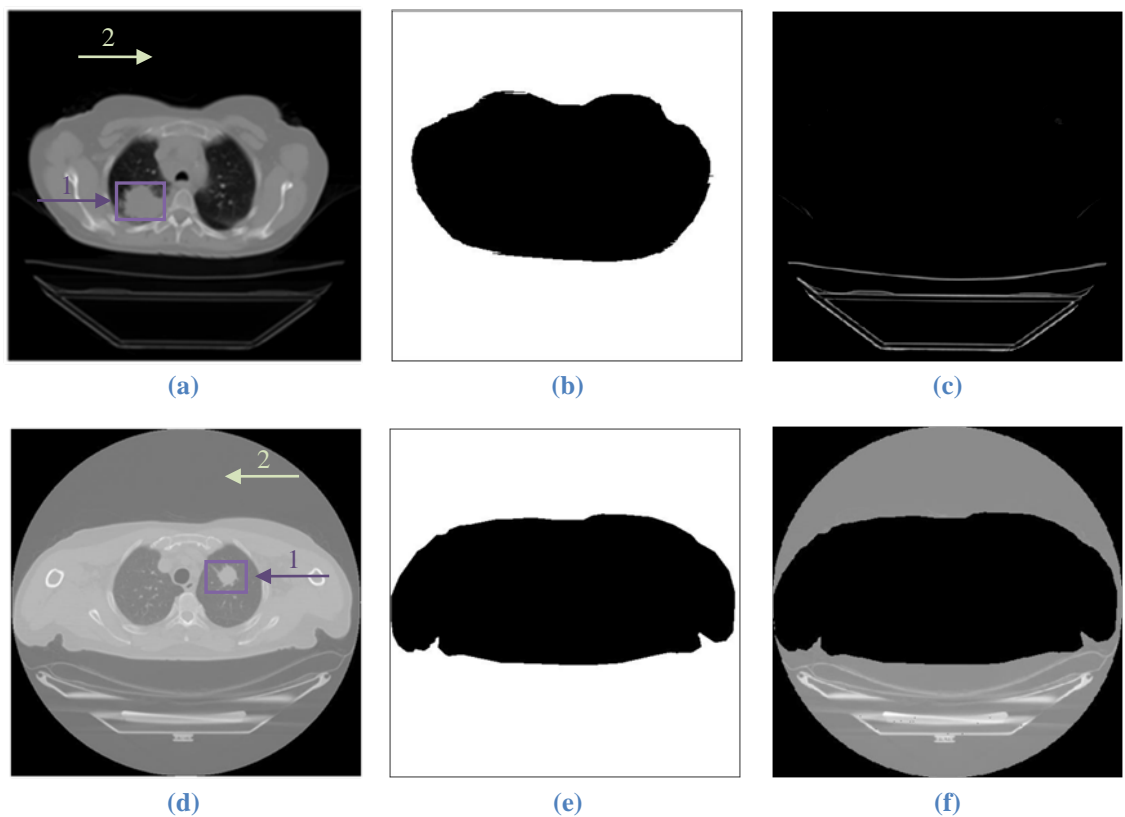

**Fig. 1.** Two images selected from a CE CT (case 1) and a NCE CT (case 5) data-sets are shown in (a) and (d), respectively. Arrow 1 shows the selected lung tumour area, and the selected open-air region – indicated by arrow 2 – which is used for noise estimation is shown in (c) and (f) after excluding the patients' body by the masks in (b) and (e).

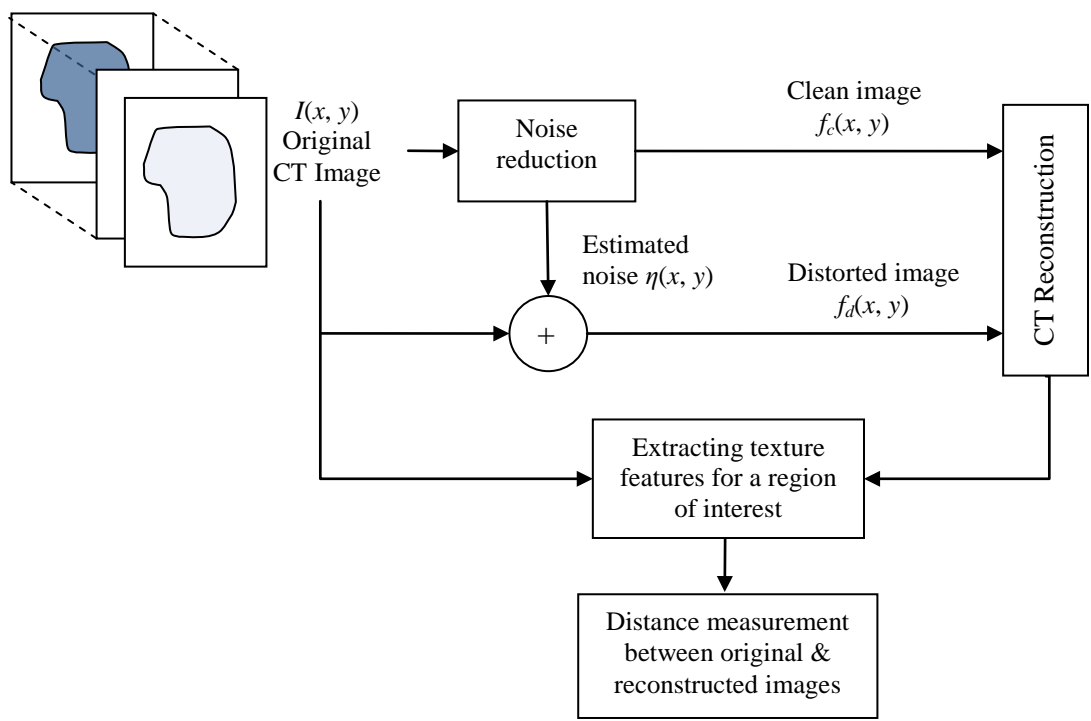

**Fig. 2.** Methodology used to assess texture measures' susceptibility to noise for lung tumour CT images.

*2.2.1 Noise estimation*

Noise was detected by examining the constant grey level area in the CT image and checking its uniformity. The background open-air region in the scanning gantry, after manually masking out the patients' body in the CT images of the CE and NCE data-sets, was chosen for analysis (see Fig. 1), and the histogram was plotted for each. Then the mean ($\mu$) and variance ($\sigma^2$) which were estimated from the plotted histogram are used to determine the parameters of three other types of noise probability density functions (PDFs) for their histograms to be plotted as well (see Table I). The selected noise types for this study were Gaussian, Rayleigh and Erlang [20]. Then the estimated histogram from the CT image will be matched against the generated noise PDFs to see to which one it best corresponds. This process is carried out for all 67 images (11 for the CE and 56 for the NCE data-sets).

The intensities histograms obtained from the CE CT uniform areas had a shape resembling additive Gaussian and multiplicative Rayleigh noise PDFs with $\mu$ and $\sigma^2$ varying between 13.2 to 17.4 and 24.7 to 65.9; respectively. While the most dominant noise in the NCE CT data-set was the Gaussian, with $\mu$ and $\sigma^2$ varying between 7.2 to 25.1 and 7.5 to 86.8; respectively. Matusita distance [21] ─ also known as first order Hellinger distance ─ which is invariant to scale in between two probability density distributions was used to compare between the original noise ($P_O$) and the three generated noise ($P_N$) distributions to see to which the measured noise is least deviated as shown in (1).

$$M(P_O, P_N) = \sqrt{\sum_i (\sqrt{P_O}(i) - \sqrt{P_N}(i))^2} \qquad (1)$$

Fig. 3 shows a histogram of noise obtained from one of the CT images compared to three different types of generated noise (Gaussian, Rayleigh and Erlang) using the estimated $\mu$ and $\sigma^2$. We can see for this case that the shape of the Rayleigh noise appears to resemble the CT noise histogram, and the distance measure supports this conclusion (see case 3 in Table II). Also in Table II, six of the examined cases showed a Rayleigh noise distribution while the rest appeared to have a Gaussian distribution. It was shown that if the variance of the estimated noise is far less than the mean intensity, the noise will approach a Gaussian distribution, whilst if it is far greater than the mean intensity will give a Rayleigh distribution [22]. Additionally, the NCE data-set in Table III shows that the noise in 51 of the 56 cases was Gaussian, while 2 was Rayleigh and 3 of Erlang type.

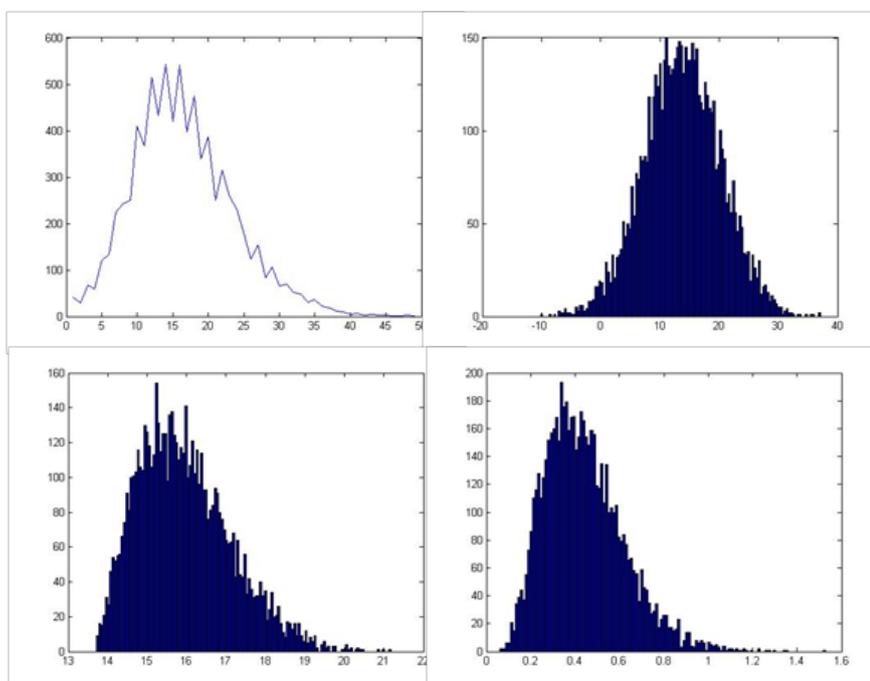

Fig. 3. From left to right and from top to bottom, histograms with $\mu_n = 13.6977$, $\sigma^2_n = 41.1472$ of transverse section of scanning table in CT images followed by corresponding generated Gaussian, Rayleigh and Erlang noises; respectively.

*2.2.2 Adaptive filtering*

Having identified the type of noise, we need to clean each of the CT images given the corresponding noise variance. As the tumour area is relatively small as compared to the total image size, an adaptive filter is needed which can reduces local noise and preserves the edges and fine structures in the CT image for subsequent accurate analysis. Since the main focus of this paper is to compare the extracted texture features robustness before and after noise reduction, thus a simple noise filter was used. An adaptive filter ($S_{xy}$) of size 5 x 5 which covers nearly 1% of the image in each step is applied for local noise reduction. Its behaviour changes adaptively depending on the statistical characteristics of the region inside the filter as defined in the following formula [20]:

$$f_c(x, y) = I(x, y) - \frac{\sigma_\eta^2}{\sigma_L^2}(I(x, y) - \mu_L) \quad (2)$$

Here $I(x, y)$ is the value of the original image suspected to have subtle noise at $(x, y)$; $\sigma^2_\eta$ the variance of the noise corrupting $f_c(x, y)$ to form $I(x, y)$; $\mu_L$ is the local mean of the pixels in $S_{xy}$; and $\sigma^2_L$, the local variance of the pixels in $S_{xy}$. In case of noise absence (i.e. $\sigma^2_\eta = 0$) the filter will return the original image. Also it preserves the edges in case the local variance is high. If noise and local variances are equal the filter return the arithmetic mean value of the pixels in $S_{xy}$.

In order to study the impact of increased noise on texture analysis measures used in CT images, a distorted image $f_d(x, y)$ is generated by simply adding the estimated noise $\eta(x, y)$ — which is a by-product of the adaptive filtering process — to the original image $I(x, y)$, as in (3).

$$f_d(x, y) = I(x, y) + \eta(x, y) \quad (3)$$

2.3 CT image reconstruction

An open-source software called CTSim [23] was used in the simulation process to reconstruct the CT images. The software simulates the process of collecting X-ray data of phantom objects. The intensity of each pixel in the original DICOM CT image was considered as a rectangle object of unit distance representing the X-ray attenuation coefficient referring to that position. By the end of this stage, three different CT images represent each case, which are the original and two versions acquired under different conditions. The amount of estimated subtle noise represented by the difference between the clean and original NCE CT image of Fig.1d is shown in Fig.4b. Also a horizontal profile along the middle of a 32 x 32 pixels background ROI indicated by arrow 2 in Fig. 1a illustrates the difference between the original, clean and noisy CT image versions is shown in Fig. 5. Texture analysis is then performed first on the 33 CE CT images in the CE data-set and then on the 168 NCE CT data-set as described in the next section.

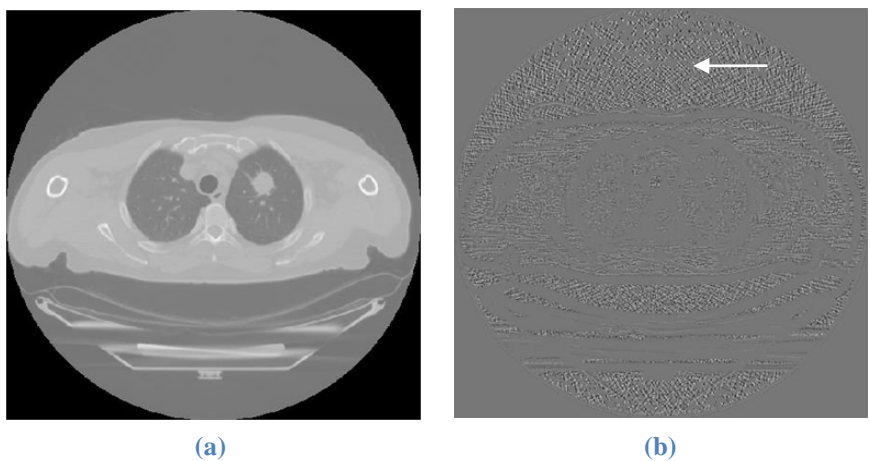

(a)          (b)

**Fig. 4. Noise suppression after adaptive filtering (a) is the clean reconstructed CT image, and (b) is the difference image between the clean and original image. The amount of reduced subtle noise becomes obvious in the open-air background above the patient which is indicated by an arrow.**

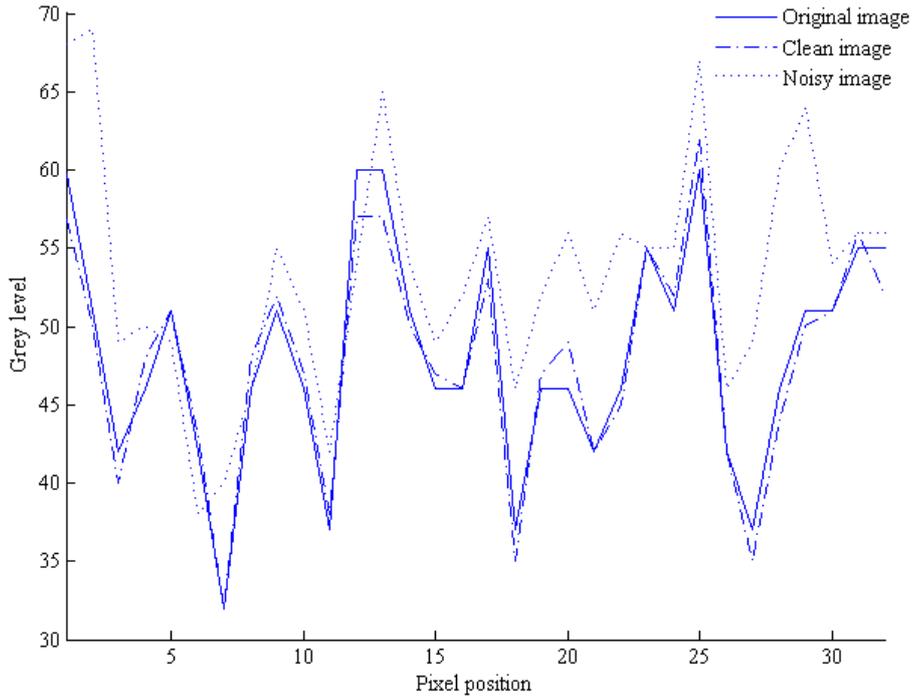

**Fig. 5.** One dimensional horizontal gray-level profile along the middle of an extracted ROI from a background (open-air) uniform area in the CT image of Fig.1a and its corresponding reconstructed clean and noisy versions.

2.4 Texture feature extraction

As different lung tumours vary in size, which relies upon the stage of development and aggression, a size that ensures capturing of the texture variation in each ROI is needed. Smaller areas would not have sufficient pixels to reliably compute the texture parameters, while larger areas would exclude relatively small size tumours from calculations. Therefore, we have empirically chosen an ROI of size 32 x 32 pixels to be extracted from each tumour region of the 201 CT images representing both data-sets (33 CE and 168 NCE), as this chosen size would balance the trade-off between tumour size and texture area. Seven different texture analyses methods were applied to analyse the texture characteristics of the ROIs. These methods are represented by Gaussian Markov random field (GMRF) and fractal dimension (FD) which are model based, and autocovariance function (ACF), runlength matrix (RLM) and grey level co-occurrence matrix (CM) which are statistical based, and discrete wavelet packet (WP) transform and Gabor filters (GF) being wavelet-based.

*2.4.1 Model-based features methods*

*Gaussian Markov random fields*

Based upon the Markovian property, which is simply the dependence of each pixel in the image on its neighbours only, a Gaussian Markov random field model (GMRF) for third order Markov neighbours was used [24] (see Fig. 6). Seven GMRF parameters were estimated using least square error estimation method.

The GMRF model is defined by the following formula:

$$p(I_{xy}|I_{kl},(k,l) \in N_{xy}) = \frac{1}{\sqrt{2\pi\sigma^2}} \exp\left\{\frac{1}{2\sigma^2}\left(I_{xy} - \sum_{l=1}^{n} \alpha_l s_{xy;l}\right)^2\right\} \quad (4)$$

where the right hand side of (4) represents the probability of a pixel ($x$, $y$) having a specific grey value $I_{xy}$ given the values of its neighbours, $n$ is the total number of pixels in the neighbourhood $N_{xy}$ of pixel $I_{xy}$, which influence its value, $\alpha_l$ is the parameter with which a neighbour influences the value of ($x$, $y$), and $s_{xy;l}$ is the sum of the values of the two pixels which are in symmetric position about ($x$, $y$) and which influence the value of ($x$, $y$) with identical parameters where,

$$s_{xy;1} = I_{x-1,y} + I_{x+1,y} \quad s_{xy;3} = I_{x-2,y} + I_{x+2,y} \quad s_{xy;5} = I_{x-1,y-1} + I_{x+1,y+1}$$
$$s_{xy;2} = I_{x,y-1} + I_{x,y+1} \quad s_{xy;4} = I_{x,y-2} + I_{x,y+2} \quad s_{xy;6} = I_{x,y-2} + I_{x+1,y-1}$$

For an image of size *M* and *N* the GMRF parameters *α* and *σ* are estimated using least square error estimation method, as follows:

$$\begin{pmatrix} \alpha_1 \\ \vdots \\ \alpha_n \end{pmatrix} = \left\{ \sum_{xy} \begin{bmatrix} s_{xy;1}s_{xy;1} & \cdots & s_{xy;1}s_{xy;n} \\ \vdots & \ddots & \vdots \\ s_{xy;n}s_{xy;1} & \cdots & s_{xy;n}s_{xy;n} \end{bmatrix} \right\}^{-1} \sum_{xy} I_{xy} \begin{pmatrix} s_{xy;1} \\ \vdots \\ s_{xy;n} \end{pmatrix} \quad (5)$$

$$\sigma^2 = \frac{1}{(M-2)(N-2)} \sum_{xy} \left[ I_{xy} - \sum_{l=1}^{n} \alpha_l s_{xy;l} \right]^2 \quad (6)$$

| $I_{x+2,y-2}$ | $I_{x+2,y-1}$ | $I_{x+2,y}$ | $I_{x+2,y+1}$ | $I_{x+2,y+2}$ |
|---|---|---|---|---|
| $I_{x+1,y-2}$ | $I_{x+1,y-1}$ | $I_{x+1,y}$ | $I_{x+1,y+1}$ | $I_{x+1,y+2}$ |
| $I_{x,y-2}$ | $I_{x,y-1}$ | $I_{xy}$ | $I_{x,y+1}$ | $I_{x,y+2}$ |
| $I_{x-1,y-2}$ | $I_{x-1,y-1}$ | $I_{x-1,y}$ | $I_{x-1,y+1}$ | $I_{x-1,y+2}$ |
| $I_{x-2,y-2}$ | $I_{x-2,y-1}$ | $I_{x-2,y}$ | $I_{x-2,y+1}$ | $I_{x-2,y+2}$ |

**Fig. 6**. **Third order Markov neighbourhood (in dark) for a sample image pixel $I_{xy}$.**

*Fractal dimension*

Fractals are used to describe non-Euclidean structures that show self-similarity at different scales. There are several fractal models used to estimate the fractal dimension; the fractal Brownian motion which is the mean absolute difference of pixel pairs as a function of scale as shown in (7) was adopted [25].

$$E(\Delta I) = K \Delta r^H \quad (7)$$

Herein, $\Delta I = |I(x_2, y_2) - I(x_1, y_1)|$ is the mean absolute difference of pixel pairs; $\Delta r = [(x_2 - x_1) + (y_2 - y_1)]^{1/2}$ is the pixel pair distances; *H* is called the Hurst coefficient; and *K* is a constant. The fractal dimension (FD) can be then estimated by plotting both sides of (7) on a log-log scale and *H* will represent the slope of the curve that is used to estimate the FD as: FD = 3 – H. By operating pixel by pixel, an FD image was generated for each ROI where each pixel has its own FD value. Then first order statistical features were derived: mean, variance, kurtosis, lacunarity, and skewness.

*2.4.2 Statistical-based features methods*

*Co-occurrence matrices*

The grey level co-occurrence matrix (CM) $P_{CM}(x, y \mid \delta, \theta)$ represents the joint probability of certain sets of pixels having certain grey-level values. It calculates how many times a pixel with grey-level $x$ occurs jointly with another pixel having a grey value $y$. For an $M \times N$ image, where $M = N$, and by varying the displacement vector $\delta$ between each pair of pixels, up to $M - 1$ CMs with different directions $\theta$ can be generated. The CM can be formally defined as [24]:

$$P_{CM}(x, y \mid \delta, \theta) = \sum_m \sum_n \Delta(x - I(m,n))\Delta(y - I(m + \delta\cos\theta, n + \delta\sin\theta)) \quad (8)$$

where $I(m, n)$ is the image grey value of pixel $(m, n)$; $I(m + \delta\cos\theta, n + \delta\sin\theta)$ is the grey value of another pixel at distance $\delta$ and direction $\theta$; $P_{CM}(x, y \mid \delta, \theta)$ is the total number of paired pixels identified in the image with grey values $x$ and $y$. For the above expression

$$\Delta(a - b) = \begin{cases} 1 & \text{if} \quad a = b \\ 0 & \text{if} \quad a \neq b \end{cases}$$

Having the CM normalised to be represented as a joint probability density function, we can then in four different directions (0°, 45°, 90° & 135°) derive four second order statistical features which are most commonly used in the literature, which are: correlation, energy, dissimilarity and entropy.

*Run-length matrices*

Another way for extracting higher order statistical texture features is the use of grey level run-length matrix (RLM) $P_{RLM}(x, y \mid \theta)$, where it can be defined as the number of occurrence of runs with pixels of gray level $x$ and run length $y$ co-linear in a given direction $\theta$ [26]. Then four statistical features were derived in four directions (0°, 45°, 90° & 135°): short run emphasis, long run emphasis, gray level non-uniformity and run length non-uniformity.

*Autocovariance function*

The autocovariance function (ACF) is the autocorrelation function after subtracting the mean. It is a way to investigate non-randomness by looking for replication of certain patterns in an image. The ACF is defined as:

$$\rho(x, y) = \frac{1}{(M - x)(N - y)} \sum_{m=1}^{M-x} \sum_{n=1}^{N-y} (I(m,n) - \mu)(I(m + x, n + y) - \mu) \quad (9)$$

where $I(m, n)$ is the grey value of an $M \times N$ ROI, $\mu$ is the mean before processing and $x$, $y$ are the amount of shifts. After calculating the ACF the peaks of the horizontal and vertical margins were fitted using least squares by an exponential and parabola functions. Therefore, each ROI is represented by eight different parameters, which are the horizontal and vertical margins values referring to the ACF, and the exponential and parabola fittings.

*2.4.3 Wavelet-based features methods*

*Gabor filters*

The Gabor filter (GF) is a Gaussian modulated sinusoidal with a capability of multi-resolution decomposition due to its localization in the spatial and spatial-frequency domain. A dyadic Gabor filter bank covering the spatial-frequency domain with multiple orientations was applied [27]. The real impulse response of a 2-D sinusoidal plane wave with orientation $\theta$ and radial centre frequency $f_0$ modulated by a Gaussian envelope with standard deviations $\sigma_x$ and $\sigma_y$ is given by

$$h(x, y) = \exp\left\{-\frac{1}{2}\left[\frac{x^2}{\sigma_x^2} + \frac{y^2}{\sigma_y^2}\right]\right\} \cos(2\pi f_0 x) \quad (10)$$

where $x = x\cos\theta + y\sin\theta$
$y = -x\sin\theta + y\cos\theta$

Figure 7 shows the frequency response of the dyadic filter bank in the spatial-frequency domain. Having a relatively small ROI size, three radial frequencies ($2^2\sqrt{2}, 2^3\sqrt{2}$ and $2^4\sqrt{2}$) with four orientations (0°, 45°, 90° & 135°) were adopted. Finally the extracted features would represent the energy of each magnitude response.

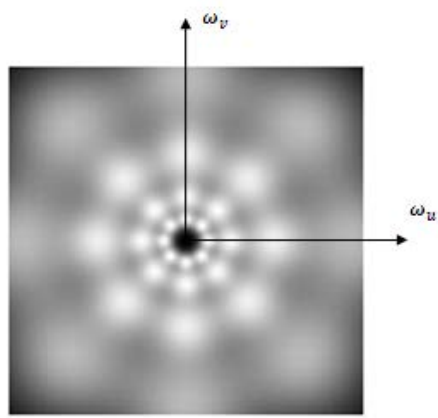

Fig. 7. Gabor filter defined in the spatial-frequency domain with 45° orientation separation.

*Wavelet packet transform*

Wavelet packet transform (WP) is a generalisation of the classical wavelet tree decomposition, providing an effective representation of the time-frequency properties [28]. In this work, each tumour ROI is decomposed down to two levels of resolution and the strongest energy subband from the leaves of each of the first level nodes (i.e. $LL_{11}$, $LH_{12}$, $HL_{13}$ and $HH_{14}$) are solely included in the ROI feature vector. An example is illustrated in Fig. 8, where a tumour ROI of the 56th case in the NCE CT data-set is extracted and decomposed into WPs down to the second level, with the corresponding subbands energy values shown in Fig.9.

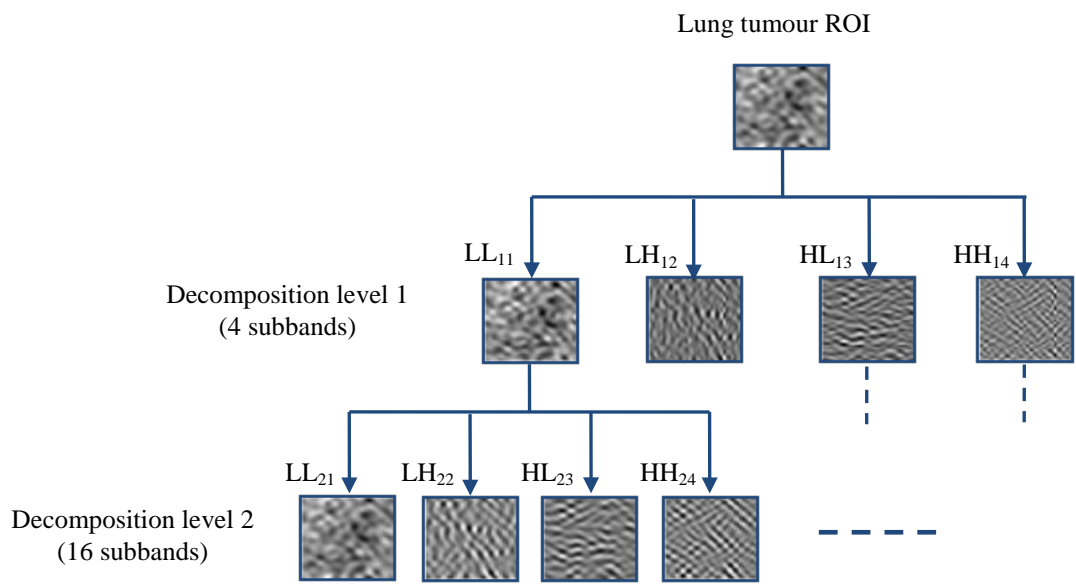

Fig. 8. A two level wavelet packet transform decomposition for an extracted lung tumor ROI of case 56 in the NCE CT data-set. The subscripts for each subband (e.g. $HL_{23}$) indicate the level and the subband number; respectively.

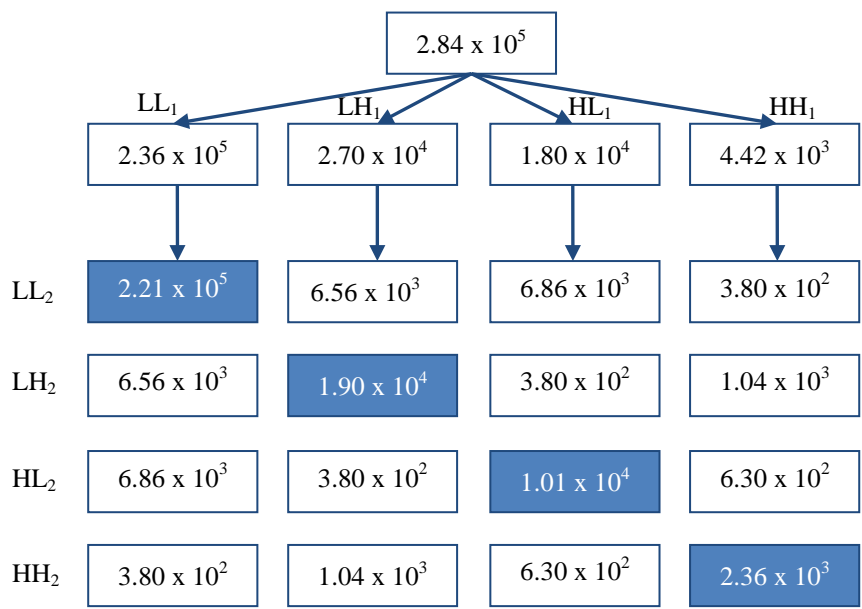

Fig. 9. Subbands' energy values for the tumour ROI decomposition in Fig. 8. The highlighted subbands represent the highest energies in the second level of decomposition.

2.5 Measuring separability quality

The final phase in this paper is the comparison process where the reconstructed images are compared to the original CT images in terms of how much deviation is incurred in the reconstructed images due to noise (removal/addition) after normalising all extracted texture measures. The Fisher's criterion which is a nonparametric statistical distance measure was used for comparison by assessing the quality of separability of two classes. It represents the ratio of the between-class variance relative to the within-class variance. In case of a multi-feature vector, the distance can be measured by the following formula [29]:

$$J(W_j) = \frac{W_j^T S_B W_j}{W_j^T S_W W_j} \qquad (11)$$

where $W_j$, $j = 1, 2, \ldots k$, are the set of discriminant vectors; $S_B$ and $S_W$ are the between-class and within-class scatter matrices. Although distance measures are often used in determining accuracy of clusters separability, it is used here to indicate how non-separable (i.e. close) the reconstructed images are to the original. Our aim is to find the best non-separable texture measure between the original and reconstructed images which is less susceptible to noise. For our case smaller values show better performance since the larger the Fisher criterion values the more significant the difference between the two assessed classes.

3. Results and Discussion

For both CE and NCE CT data-sets, the class separability between the original CT image and its reconstructed clean and distorted versions measured by Fisher's distance is listed in Table IV by $J_{oc}$ and $J_{on}$; respectively. In order to make the analysis of Table IV easier, the texture features are sorted in ascending order in Table V, placing the least separable at the top and vice versa. From the first glance, it can be seen that the WP was the least affected by noise in both data-sets and for the cleaned and distorted versions as well. Regarding the CE CT data-set, there was no difference in the order between the used texture methods for characterising the clean and distorted versions CT images. This shows that CE CT images can assist in highlighting the lung texture variations, and hence reducing the effect of distortion on the extracted tissue characteristics.

As the noise found in the both data-sets was subtle, the distorted CT image reconstructed versions contributed in giving some emphasis to this noise. However, we assume that the uncorrelated noise encountered in the CT images is within acceptable ranges and not quite visible to the extent that would deform the structure of the observed ROI ─ which is in fact the case in most captured CT images. Additionally, the FD, ACF and GF, which came next to the WP, scored nearly a similar score in terms of separability, with the FD being the least affected by noise amongst them. The RLM, CM and GMRF did not perform as efficient as the rest of the previously mentioned texture methods, with the GMRF scoring the most susceptible to noisy conditions. On the other hand, the NCE CT data-set witnessed some change in the order of the tested texture methods, and in-between the clean and distorted CT versions. Although CE CT images would improve the reliability of the used texture methods for analysis, these images are not easy to acquire in comparison to the conventional NCE CT images, where patients need to be injected with a special contrast agent into a large blood vessel prior to image capturing, rendering it unpleasant for patients.

It should be noted that this work does not intend to compare the performance of these texture measures in terms of discrimination capability or which provides a better characterisation of the lung tissue, but to assess their immunity when used under the presence and absence of the same noise detected in the investigated CT images. However, it is no coincidence for WP and FD texture measures which were the least susceptible to noise to offer good performance in lung tissue analysis [8, 11]. Thus, a relation between how immune is a texture measure to noise and how effectively it can characterise a texture under investigation exist; especially when analysing tissue texture which is considered non-stationary; creating another challenge besides noise. The improved capability of the WP and FD in analysing lung tissue could be interpreted as their ability to exclude noise with minimum effect on the analysed texture. That is, by decomposing the ROI into several subbands for the WP case, high subbands can be easily eliminated from further decomposition where random noise is usually present in the high subbands. Herein, the subbands with the strongest energy were only selected from the leaves of each decomposed subband for the feature vector. Also the FD can mitigate the effect of noise as it gives a quantitative assessment of the roughness of the surface by examining the texture ROI at different scales, thus the noise would not have a similar affect at all scales. Another point is that the ACF came second for the $J_{oc}$ NCE CT images and third in the rest of Table V, nevertheless it has a poor performance in image classification [30]. This means that although the ACF is less susceptible to noise, yet this is due to its initial poor characterisation of texture resulting in the little difference between the original and the clean and distorted image versions.

Observing the number of features generated from each texture measure, it seems that the number of extracted features plays an important role in susceptibility to noise. CM or RLM which extracts 16 different features was more prone to noise as compared to the WP which had only 4 features. This might be due to the fact that texture measures with large number of features tend to capture more variations of the intensity, and as a result the probability of noise contribution would be amplified. Furthermore, not all extracted features relevant to a specific texture method have the same discriminating power, and thus optimisation might result in a fewer number of features which can still efficiently characterise the selected ROI and be less susceptible to noise distortion. On the other hand, although some studies reported signal dependent Gaussian noise distributions in low dose CT images [31], and the Gaussian was the dominant type of noise in the NCE CT data-set, this work showed that other types of noise rather than Gaussian can be encountered even when using the same CT scanner, which might be due to external distortion (i.e. not related to the CT scanner) when some of the images were acquired.

This indicates that noise can have some impact on the variability of diagnosis reports depending on the used texture measure for analysis and classification. Some texture measure are more reliable in terms of classification [30, 32], yet their accuracy might start to give misleading results in case of noise presence, causing an increase in inaccuracy as noise becomes more obvious. Therefore, accuracy and noise susceptibility must be taken into consideration by the physician depending on the type of analysis and the area of texture. Given the variation in size, shape and stage between the different extracted lung tumours in this study, the texture measures were applied to a 32 x 32 size ROI which insures the inclusion of all tumours (i.e. the small size ones as well). So based on the results and whenever it becomes difficult to extract a sufficiently large ROI analysis, physicians can use texture measures which exhibited the least susceptibility to noise such as WP or FD for small areas (e.g. size $\leq$ 32 x 32) of texture where the probability of noise deforming the structure of the texture is higher, and use the texture measures known for their good capability in texture discrimination but there performance was more prone to noise, as the CM or GMRF for example, for larger ROIs. Also, the Fisher's distance showed that the five of the seven clean CE CT reconstructions are much nearer to the original from the distorted ones, therefore adaptive filtering can assist in improving some of the texture measures' efficiency.

Possible improvement in order to enhance the reliability of the reported results in this work is that noise susceptibility comparison can be made after applying an optimum feature selection technique for each of texture methods. By that, features with weak discriminating capability are eliminated and the total number of features for each method is reduced to a minimum. Also the difference in the acquired images slice thickness between the CE and the NCE data-sets might affect the accuracies of the measured Fisher's distances; yet this needs to be further investigated.

A future trend would be assessing the quality of the extracted features under reduced radiation dose (RRD) CT images. For patient's safety and to avoid the relatively high dose of radiation in CT modalities, RRD CT images are acquired in case of children as their tissue is more sensitive to the radiation effect [33, 34], or for adults depending on the kind of organ under investigation. Lowering the radiation dose can be done by decreasing the tube current mAs or beam energy kV with high pitches or table-speed, and using thicker slices [35]. The effect of RRD CT images, which usually yield noisier and lower contrast images, on extracted texture measures can assist in performance comparison under noisy and lower quality image conditions. Also investigating the texture measures susceptibility under correlated noise, and in other modalities when other types of noise might be present would be advantageous.

**4. Conclusion**

The robustness or how well a specific texture measure can tolerate noise in a CT image of lung tumour texture was presented. Susceptibility of seven different texture analysis measures to noise was investigated by using Fisher's distance to compare the original CT images with their corresponding reconstructed clean and noisy versions. From the two different used data-sets, Rayleigh and Gaussian noise was encountered in the CE data-set, while the Gaussian noise was the dominant in the NCE data-set. It was shown that CE CT images yields more stable results in comparison to its conventional (NCE) counterparts, while the WT and GF wavelet-based texture methods being stable in both data-sets. The WP and FD which could characterise the lung tissue better than the other texture measures were the least effected by noise. Moreover, WP and FD had the least number of extracted features in comparison to RLM and CM which had the highest number of features, and the last two with the GMRF were most sensitive to noise. Finally, it was shown that adaptive filtering can assist in the reduction of subtle noise, and hence offer better texture accuracy.

**Acknowledgment**

Thanks to the Clinical Imaging Science Centre at the University of Sussex for provision of the CT image data-sets used in this study. Also the author would like to thank the University of Jordan, Amman, Jordan, for funding and supporting this research, and the anonymous reviewers for their constructive comments.